\documentclass[journal]{IEEEtran}
\usepackage{cite}
\usepackage[utf8]{inputenc}
\usepackage{amsfonts}
\usepackage[colorinlistoftodos]{todonotes}
\usepackage{algpseudocode}
\usepackage{times}
\usepackage{epsfig}
\usepackage{amsmath}
\usepackage{amssymb} 
\usepackage{algorithm}
\usepackage{multirow} 
\usepackage{bibentry}
\usepackage{graphicx}
\usepackage{url}
\usepackage{multirow}
\usepackage{booktabs}
\usepackage{enumerate}
\usepackage{textcomp}
\usepackage{subfigure}
\usepackage{upgreek}
\usepackage{ stmaryrd }

\ifCLASSINFOpdf
\else
\fi
\begin{document}
\title{Minimum Margin Loss for Deep Face Recognition}

\author{Xin~Wei,~\IEEEmembership{Student Member,~IEEE,}
Hui~Wang,~\IEEEmembership{Member,~IEEE,}
Bryan~Scotney,
and~Huan~Wan% <-this % stops a space
\thanks{Xin Wei is with the School of Computing, Ulster University at Jordanstown, BT370QB UK (e-mail: wei-x@ulster.ac.uk).}% <-this % stops a space
\thanks{Hui Wang is with the School of Computing, Ulster University at Jordanstown, BT370QB UK (e-mail: h.wang@ulster.ac.uk).}% <-this % stops a space
\thanks{Bryan Scotney is with the School of Computing, Ulster University at Coleraine, BT521SA UK (e-mail: bw.scotney@ulster.ac.uk).}% <-this % stops a space
\thanks{Huan Wan is with the School of Computing, Ulster University at Jordanstown, BT370QB UK (e-mail: wan-h@ulster.ac.uk).}}% <-this % stops a space

% ^------------^------------^----Do not want these spaces!

\markboth{Journal of \LaTeX\ Class Files,~Vol.~14, No.~8, August~2015}%
{Shell \MakeLowercase{\textit{et al.}}: Bare Demo of IEEEtran.cls for IEEE Journals}
% *** name in the headers of peer review papers. ***



\maketitle

\begin{abstract}

Face recognition has achieved great progress owing to the fast development of the deep neural network in the past a few years. As an important part of deep neural networks, a number of the loss functions have been proposed which significantly improve the state-of-the-art methods. In this paper, we proposed a new loss function called Minimum Margin Loss (MML) which aims at enlarging the margin of those overclose class centre pairs so as to enhance the discriminative ability of the deep features. MML supervises the training process together with the Softmax Loss and the Centre Loss, and also makes up the defect of Softmax + Centre Loss. The experimental results on MegaFace, LFW and YTF datasets show that the proposed method achieves the state-of-the-art performance, which demonstrates the effectiveness of the proposed MML. 

\end{abstract}


% \fi
\IEEEpeerreviewmaketitle

\section{Introduction}
In the past ten years, Deep Learning-based methods have achieved great progress in various computer vision areas, including face recognition \cite{schroff2015facenet,sun2014deep,deng2018arcface,huang2017densely}, object recognition \cite{liang_recurrent_2015,li_multistage_2016,yan2017incremental}, action recognition \cite{wang_action_2015,Veeriah_2015_ICCV,varol_long-term_2018,ma_region-sequence_2018} and so on. Among these areas, the progress on face recognition is particularly remarkable because of the development of two important aspects -- larger face datasets and better loss functions.

As a crucial factor, the scale and the quality of the training datasets directly influence the performance of a DNN model. Currently, there are a few public available large-scale face datasets, for example, MS-Celeb-1M \cite{guo2016msceleb}, VGGFace2 \cite{cao2017vggface2}, MegaFace \cite{kemelmacher2016megaface} and CASIA WebFace \cite{yi2014learning}. As shown in Table \ref{Large-scale datasets}, CASIA WebFace consists of 0.5M face images; VGGFace2 contains totally 3M face images but only from 9K identities; while MS-Celeb-1M and MegaFace contain many more identities and more images, which have bigger potential for training a better DNN model. However, both MS-Celeb-1M and MegaFace have the problem of long-tailed distribution \cite{zhang_range_2017}, which means a minority of people owns a majority of face images while a large number of people have very limited face images. Using the datasets with long-tailed distribution, the trained model tends to overfit the classes with rich samples and weaken the generalisation ability on the long-tailed portion \cite{zhang_range_2017}. Specifically, in order to separate different classes, the classes with rich samples tend to have a relatively larger margin between their class centres; conversely, as the classes with limited samples only occupy a small space and are easy to be compressed and separated, they tend to own a relatively smaller margin between their class centres. Thus, it is reasonable to consider setting a minimum margin to rectify this bias. In this paper, we will focus on the overclose pairs of class centre and propose our loss based on the minimum margin.

% \usepackage{booktabs}
\begin{table}[]
\centering
\caption{Statistics for recent public available large-scale face datasets.}
\label{Large-scale datasets}
\begin{tabular}{@{}ccccc@{}}
\toprule
& MS-Celeb-1M & VGGFace2 & MegaFace & CASIA \\ \midrule
\#Identities & 100K & 9K & 672K & 11K \\
\#Images & 10M & 3M & 5M & 0.5M \\
Avg per Person & 105 & 323 & 7 & 47 \\ \bottomrule
\end{tabular}
\end{table}

Besides the training set, another important aspect is the loss function which directs the networks to optimise their weights during the training process. Nowadays, the best performing loss functions can be roughly divided into two types \cite{deng_marginal_2017, wang_additive_2018, deng_arcface:_2018}: the loss functions based on Euclidean distance and the loss functions based on Cosine distance. Most of them are derived from Softmax Loss by adding a penalty or modifying softmax directly.

The loss functions based on Euclidean distance include Contrastive Loss \cite{sun2014deep}, Triplet Loss \cite{schroff2015facenet}, Centre Loss \cite{wen_discriminative_2016}, Range Loss \cite{zhang_range_2017}, Marginal Loss \cite{deng_marginal_2017} and so on. These methods aim at improving the discriminative ability of features by maximising the inter-class distance or minimising the intra-class distance. Contrastive Loss inputs the networks with two types of sample pair -- the positive sample pair (two faces from the positive class) and the negative sample pair (two face images from the negative class). Contrastive Loss minimises the Euclidean distance of the positive pairs and penalises the negative pairs that have a distance smaller than the threshold. Triplet Loss uses the triplet as the input which includes a positive sample, a negative sample and an anchor. An anchor is also a positive sample, which is initially closer to some negative samples than it is to some positive samples. During the training, the anchor-positive pairs are pulled together while the anchor-negative pairs are pushed apart as much as possible. However, the selection of the sample pairs and the triplets is laborious and time-consuming for both Contrastive Loss and Triplet Loss. Centre Loss, Marginal Loss and Range Loss add another penalty to implement the joint supervision with Softmax Loss. Specifically, Centre Loss adds a penalty to Softmax by calculating and restricting the distances between the within-class samples and the corresponding class centre. Marginal Loss considers all the sample pairs in a batch and forces the sample pairs from the different classes to have a margin larger than the threshold $\theta$ while forcing the samples from the same classes to have a margin smaller than the threshold $\theta$. But it is overstrict to force the two farthest samples in a class to have a distance smaller than the two nearest samples from the different classes, which makes the training procedure hard to converge. Range Loss calculates the distances of the samples within each class, and chooses two sample pairs which have the largest distances as the intra-class constraint; simultaneously, Range Loss calculates the distance of each class centre pair, and forces the class centre pair that has the smallest distance to have a larger margin than the designated threshold. However, only considering one centre pair each time is not comprehensive, as more centre pairs may have the margins smaller than the designated threshold and the training procedure is hard to completely converge because of the slow learning speed.

The loss functions based on on Cosine distance include L-Softmax Loss \cite{liu_large-margin_2016}, A-Softmax Loss \cite{liu_sphereface:_2017}, AM-Softmax Loss \cite{wang_additive_2018}, ArcFace \cite{deng_arcface:_2018} and so on. L-Softmax reformulates the output of softmax layer from $W\cdot f$ to $|W|\cdot |f|\cdot cos\theta$ so as to transform the Euclidean distance to Cosine distance, and also add multiplicative angular constraints to $cos\theta$ to enlarge the angular margins between different identities. Based on L-Softmax Loss, A-Softmax applies weight normalisation, so $W\cdot f$ is further reformulated to $|f|\cdot cos\theta$ which simplifies the training target. However, after using the same multiplicative angular constraints, both L-Softmax and A-Softmax Loss are difficult to converge. So annealing optimization strategy is adopted by these two methods to help the algorithm to converge. To improve the convergence of A-Softmax, Wang et al. \cite{wang_additive_2018} propose AM-Softmax which replaces the multiplicative angular constraints with the additive angular constraints, namely, transforms $cos(m\theta)$ to $cos\theta-m$. Besides, AM-Softmax also applies feature normalisation and introduces the global scaling factor $s=30$ which makes $|W|\cdot |f|=s$. Hence, the training target $|W|\cdot |f|\cdot cos\theta$ is again simplified to $s\cdot cos\theta$. ArcFace also utilises the additive angular constraints, but it changes $cos(m\theta)$ to $cos(\theta+m)$ which makes it have better geometric interpretation. Both AM-Softmax and ArcFace adopt weight normalisation and feature normalisation which restrict all the features to lie on a hypersphere. However, is it overstrict to force all the features to lie on a hypersphere instead of a wider space? Why and how do weight normalisation and feature normalisation benefit the training procedure? These questions are difficult to answer explicitly, and some evidence shows that ``soft'' feature normalisation may lead to better results \cite{zheng_ring_2018}.

Inspired by Softmax Loss, Centre Loss and Marginal Loss, we propose the Minimum Margin Loss (MML) in this paper which aims at forcing all the class centre pairs to have a distance larger than the specified minimum margin. Different from Range Loss, MML penalises all the `unqualified' class centre pairs instead of only penalising the centre pair that has the shortest distance. MML reuses the centre positions constantly updated by Centre Loss, and directs the training process by joint supervision with Softmax Loss and Centre Loss. In this way, Softmax Loss+Centre Loss+MML achieved better performance than Softmax Loss and Softmax Loss+Centre Loss while almost has no increment in computing cost. According to our knowledge, there is no loss function which considered setting a minimum margin between the class centres. However, it is necessary to have such a constraint for rectifying the bias introduced in by imbalanced data. To prove the effectiveness of the proposed method, experiments are conducted on three public datasets -- Labeled Faces in the Wild (LFW) \cite{LFWTech}, YouTube Faces (YTF) \cite{wolf2011face} and Megaface \cite{kemelmacher2016megaface} datasets. Results show that MML achieves superior performance than softmax Loss and centre Loss, while achieves competitive results compared with the state-of-the-art methods.

\section{From Softmax Loss to Minimum Margin Loss}
\subsection{Softmax Loss and Centre Loss}
Softmax Loss is the most commonly used loss function, which is presented below:
\begin{equation}\label{eq:softmax_loss}
\mathcal{L}_{S}=- \frac{1}{N} \sum_{i=1}^N log \frac{ e^{ W_{y_i}^{T}f_i+b_{y_i} } }{ \sum_{j=1}^K e^{ W_{j}^{T}f_i+b_j}} 
\end{equation}
where $N$ is the batch size, $K$ is the class number of a batch, $f_i \in R^d$ denotes the feature of the $i$th sample belonging to the $y_i$th class, $W_j \in R^d$ denotes the $j$th column of the weight matrix $W$ in the final fully connected layer and $b_{j}$ is the bias term of the $j$th class. From Eq(\ref{eq:softmax_loss}), it can be seen that Softmax Loss is designed to obtain the cross entropy between the predicted label and the true label, which in other words means the target of Softmax Loss is only to separate the features from different classes in the training set instead of learning discriminative features. Such a target is appropriate for close-set tasks, like most application scenarios of object recognition and behaviour recognition. But the application scenarios of face recognition are open-set tasks in most cases, so the discriminative ability of features has considerable influence on the performance of a face recognition system. To enhance the discriminative ability of features, Wen et al. \cite{wen_discriminative_2016} proposed the Centre Loss to minimise the intra-class distance, as shown below:
\begin{equation}\label{eq:centre_loss}
\mathcal{L}_{C}=\frac{1}{2} \sum_{i=1}^N ||f_i-c_{y_i}||_2^2
\end{equation}
where $c_{y_i}$ denotes the class centre of the $y_i$th class. Centre Loss calculates all the distances between the class centres and within-class samples, and is used in conjunction with Softmax Loss:
\begin{small}
\begin{eqnarray}\label{eq:softmax_centre_loss}
\mathcal{L}&=&\mathcal{L}_{S}+\lambda\mathcal{L}_{C}\\
&=&- \frac{1}{N} \sum_{i=1}^N log \frac{ e^{ W_{y_i}^{T}f_i+b_{y_i} } }{ \sum_{j=1}^K e^{ W_{j}^{T}f_i+b_j}} + \frac{\lambda}{2} \sum_{i=1}^N ||f_i-c_{y_i}||_2^2
\end{eqnarray}
\end{small}where $\lambda$ is the hyper-parameter for balancing the two loss functions.

\subsection{Marginal Loss and Range Loss}
After combining the Softmax Loss with the Centre Loss, the within-class compactness is significantly enhanced. But it is not enough to only use Softmax Loss as the inter-class constraint, as it only encourages the separability of features. So Deng et al. \cite{deng_marginal_2017} proposed Marginal Loss which also takes the way of joint supervision with the Softmax Loss:
\begin{eqnarray}\label{eq:marginal_centre_loss}
\mathcal{L}&=&\mathcal{L}_{S}+\lambda\mathcal{L}_{Mar}
\end{eqnarray}
\begin{small}
\begin{eqnarray}\label{eq:marginal_loss}
\mathcal{L}_{mar}= \frac{1}{N^2-N} \sum_{i,j,i \neq j}^N \bigg( \xi -y_{ij} \bigg( \theta -\bigg|\bigg| \frac{f_i}{||f_i||} - \frac{f_j}{||f_j||} \bigg|\bigg|_2^2\bigg) \bigg)_+
\end{eqnarray}
\end{small}where $f_i$ and $f_j$ are the features of the $i$th and $j$th samples in a batch, respectively; $y_{ij} \in \{\pm1\}$ indicates whether $f_i$ and $f_j$ belong to the same class, $(u)_+$ is defined as $max(u, 0)$, $\theta$ is the threshold to separate the positive pairs and the negative pairs, and $\xi$ is the error margin besides the classification hyperplane.

Marginal Loss considers all the possible combinations of the sample pairs in a batch and specifies a threshold $\theta$ to constrain all these sample pairs including the positive pairs and the negative pairs. Marginal Loss forces the distances of the positive pairs to be close up to the threshold $\theta$ while forcing the distances of the negative pairs to be farther than the threshold $\theta$. But utilising the same threshold $\theta$ to constrain both the positive and negative pairs is not proper. Because it is often the case that the two farthest samples in a class have a distance larger than the two nearest samples of the two different but closest classes. Forcibly changing this situation will make the training procedure hard to converge. 

Similar to the aforementioned methods, the Range Loss proposed by Zhang et al. \cite{zhang_range_2017} also works with softmax Loss as the supervisory signals:
\begin{eqnarray}\label{eq:range_centre_loss}
\mathcal{L}&=&\mathcal{L}_{S}+\lambda\mathcal{L}_{R}
\end{eqnarray}

Different from Marginal Loss, Range Loss consists of two independent losses, namely $\mathcal{L}_{R_{intra}}$ and $\mathcal{L}_{R_{inter}}$ to calculate the intra-class loss and inter-class loss respectively (see Eq.(\ref{eq:range_loss})).
\begin{equation}\label{eq:range_loss}
\mathcal{L}_{R}= \alpha \mathcal{L}_{R_{intra}}+ \beta \mathcal{L}_{R_{inter}}
\end{equation}
where $\alpha$ and $\beta$ are two weights for adjusting the influence of $\mathcal{L}_{R_{intra}}$ and $\mathcal{L}_{R_{inter}}$. Mathematically, $\mathcal{L}_{R_{intra}}$ and $\mathcal{L}_{R_{inter}}$ are defined as follows:
\begin{eqnarray}\label{eq:range_intra_inter_loss}
\mathcal{L}_{R_{intra}}&=& \sum_{i \subseteq K} \mathcal{L}_{R_{intra}}^i=\sum_{i \subseteq I} \frac{n}{ \sum_{j=1}^n \frac{1}{D_{ij}}} \\
\mathcal{L}_{R_{inter}}&=&max(M-D_{Centre},0)\\
&=&max(M-|| \overline{x}_\mathcal{Q}-\overline{x}_\mathcal{R}||_2^2,0)
\end{eqnarray}
where $K$ is the class number in current batch, $D_{ij}$ is the $j$th largest distance of the sample pairs in class $i$, $D_{Centre}$ is the central distance of two nearest classes in current batch, $\overline{x}_\mathcal{Q}$ and $\overline{x}_\mathcal{R}$ denote the class centres of class ${x}_\mathcal{Q}$ and ${x}_\mathcal{R}$ which have the shortest central distance, and $M$ is the margin threshold. $\mathcal{L}_{R_{intra}}$ measures all the sample pairs in a class and select $n$ sample pairs that have the large distances to build the loss for controlling the within-class compactness. As described in \cite{deng_marginal_2017}, experiments show that $n=2$ is the best choice. $\mathcal{L}_{R_{inter}}$ aims at forcing the class centre pair that has the smallest distance to have a larger margin up to the designated threshold. But there are more centre pairs that may have distances larger than the designated threshold. It is not comprehensive enough for only considering one centre pair each time which leads the training procedure to take a long time to completely converge because of the low learning speed.

\subsection{The Proposed Minimum Margin Loss}
Inspired by Softmac Loss, Centre Loss and Marginal Loss, we propose the Minimum Margin Loss (MML) in this paper. MML is used in conjunction with Softmax Loss and Centre Loss, where Centre Loss is utilised to enhance the within-class compactness, Softmax and MML are applied for improving the between-class separability. Specifically speaking, Softmax is in charge of guaranteeing the correctness of classification while MML aims at optimising the between-class margins. The total loss is shown below:
\begin{equation}\label{eq:MML}
\mathcal{L}= \mathcal{L}_{S} +\alpha \mathcal{L}_{C}+ \beta \mathcal{L}_{M}
\end{equation}
where $\alpha$ and $\beta$ are the hyper-parameters for adjusting the impact of Centre Loss and MML.

MML specifies a threshold called Minimum Margin. By reusing the class centre positions updated by Centre Loss, MML filters all the class centre pairs based on the specified Minimum Margin. For those pairs which have distances smaller than the threshold, corresponding penalties are added into to the loss value. The detail of MML is formulated as follows:
\begin{small}
\begin{eqnarray}\label{eq:marginal_loss}
\mathcal{L}_{M}= \sum_{i,j=1}^K max(||c_{i}-c_{j}||_2^2-\mathcal{M},0)
\end{eqnarray}
\end{small}where $K$ is the class number of a batch, $c_i$ and $c_j$ denote the class centres of the $i$th and $j$th classes respectively, and $\mathcal{M}$ represents the designated minimum margin. In each training batch, the class centres are updated by Centre Loss with the following two equations:
\begin{eqnarray}\label{eq:range_intra_inter_loss}
c^{t+1}_j&=&c^t_j-\gamma\Delta c^t_j \\
\Delta c^t_j &=&\frac{\sum_{i=1}^m \delta (y_i = j) \dot (c_j - f_i)}{1 + \sum_{i=1}^m \delta (y_i = j)}
\end{eqnarray}
where $\gamma$ is the learning rate of the class centres, $t$ is and the number of iteration and $\delta(condition)$ is a conditional function. If the condition is satisfied, $\delta(condition) = 1$, otherwise $\delta(condition) = 0$.

Algorithm \ref{Learning Details} shows the basic learning steps in the CNNs with the proposed \textbf{$\mathcal{L}_{S} +\mathcal{L}_{C}+ \mathcal{L}_{M}$}.

\begin{algorithm}
\caption{Learning algorithm in the CNNs with the proposed $\mathcal{L}_{S} +\mathcal{L}_{C}+ \mathcal{L}_{M}$.}
\label{Learning Details}
\begin{description}
\item[\textbf{Input:}] \hspace{0.1cm}Training samples \{$f_i$\}, initialised parameters $\theta_C$ in convolution layers, parameters $W$ in the final fully connected layer, and initialised $n$ class centres $\{c_j|j=1,2,...,n\}$. Learning rate $\mu^t$, hyperparameters $\alpha$ and $\beta$, learning rate of the class centres $\gamma$ and the number of iteration $t \shortleftarrow 1$.
\item[\textbf{Output:}] \hspace{0.4cm}The parameters $\theta_C$.
\end{description}
\begin{algorithmic}[1]
\While {not converge}
\State Calculate the total loss by $\mathcal{L}^t=\mathcal{L}^t_{S} +\alpha \mathcal{L}^t_{C}+ \beta \mathcal{L}^t_{M}$.
\State Calculate the backpropagation error $\frac{\partial \mathcal{L}^t}{\partial f^t_i}$ for each sample $i$ by $\frac{\partial \mathcal{L}^t}{\partial f^t_i}=\frac{\partial \mathcal{L}^t_S}{\partial f^t_i}+\alpha\frac{\partial \mathcal{L}^t_C}{\partial f^t_i}+\beta\frac{\partial \mathcal{L}^t_M}{\partial f^t_i}$.
\State Update $W$ by $W^{t+1}=W^{t}-\mu^t\frac{\partial \mathcal{L}^t}{\partial W^t}=W^{t}-\mu^t\frac{\partial \mathcal{L}^t_S}{\partial W^t}$.
\State Update $c_j$ for each centre $j$ by $c^{t+1}_j=c^t_j-\gamma\Delta c^t_j$.
\State Update $\theta_C$ by $\theta^{t+1}_C=\theta^t_C-\mu^t\sum^N_i \frac{\partial \mathcal{L}^t}{\partial f^t_i} \frac{\partial f^t_i}{\partial \theta^t_C}$.
\State $t \shortleftarrow t+1$.
\EndWhile
\end{algorithmic} 
\end{algorithm}

\begin{figure*}[!htbp]
\begin{minipage}{0.32\textwidth}
\subfigure[Without using MML]{
\label{fig:h1} %% label for second subfigure
\includegraphics[width=2.4in,height=1.8in]{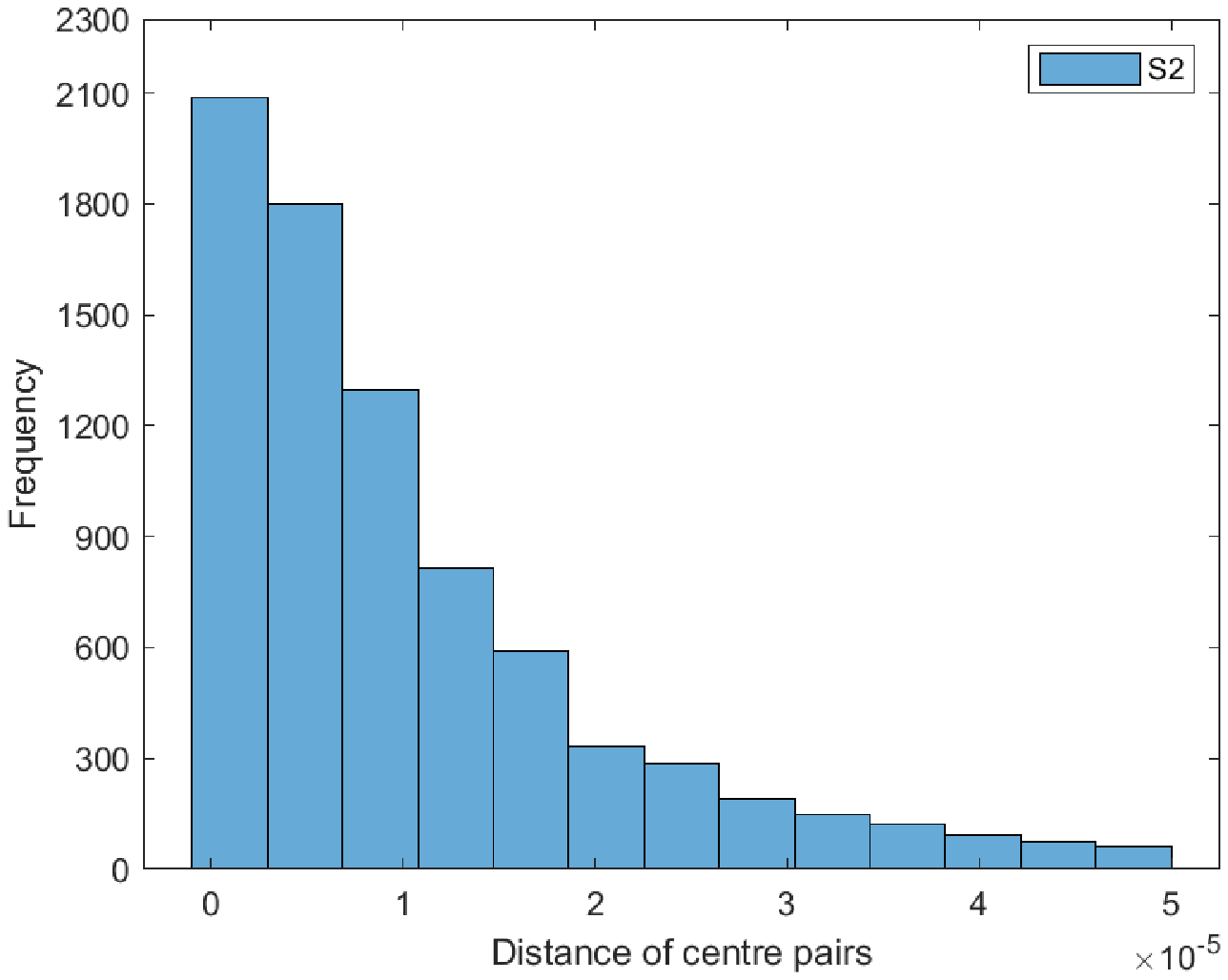}}
\end{minipage}
\begin{minipage}{0.32\textwidth} 
\hfill
\subfigure[After using MML]{
\label{fig:h2} %% label for first subfigure
\includegraphics[width=2.4in,height=1.8in]{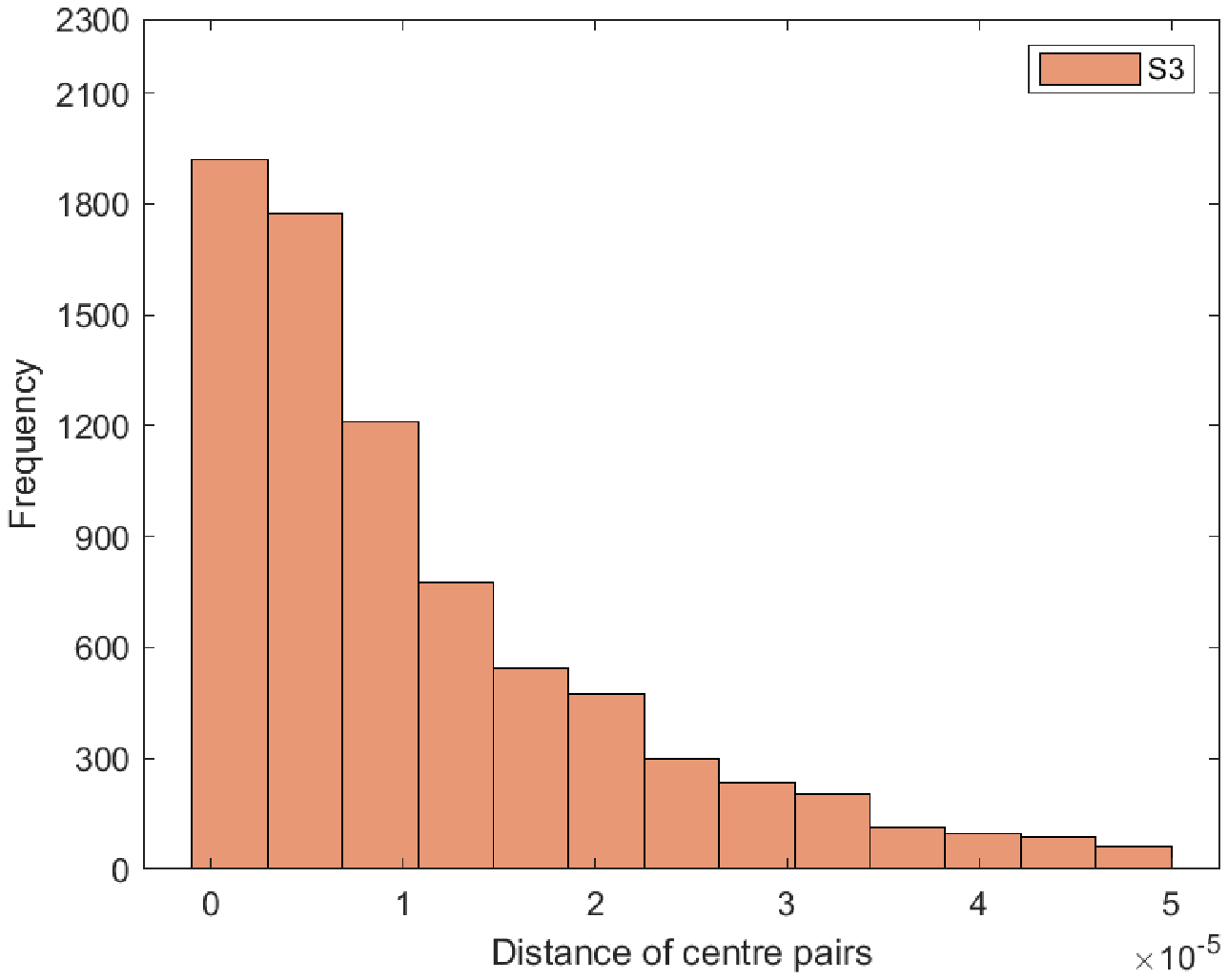}}
\end{minipage}
\begin{minipage}{0.32\textwidth} 
\hfill
\subfigure[Comparison between S2 and S3]{
\label{fig:h3} %% label for first subfigure
\includegraphics[width=2.4in,height=1.8in]{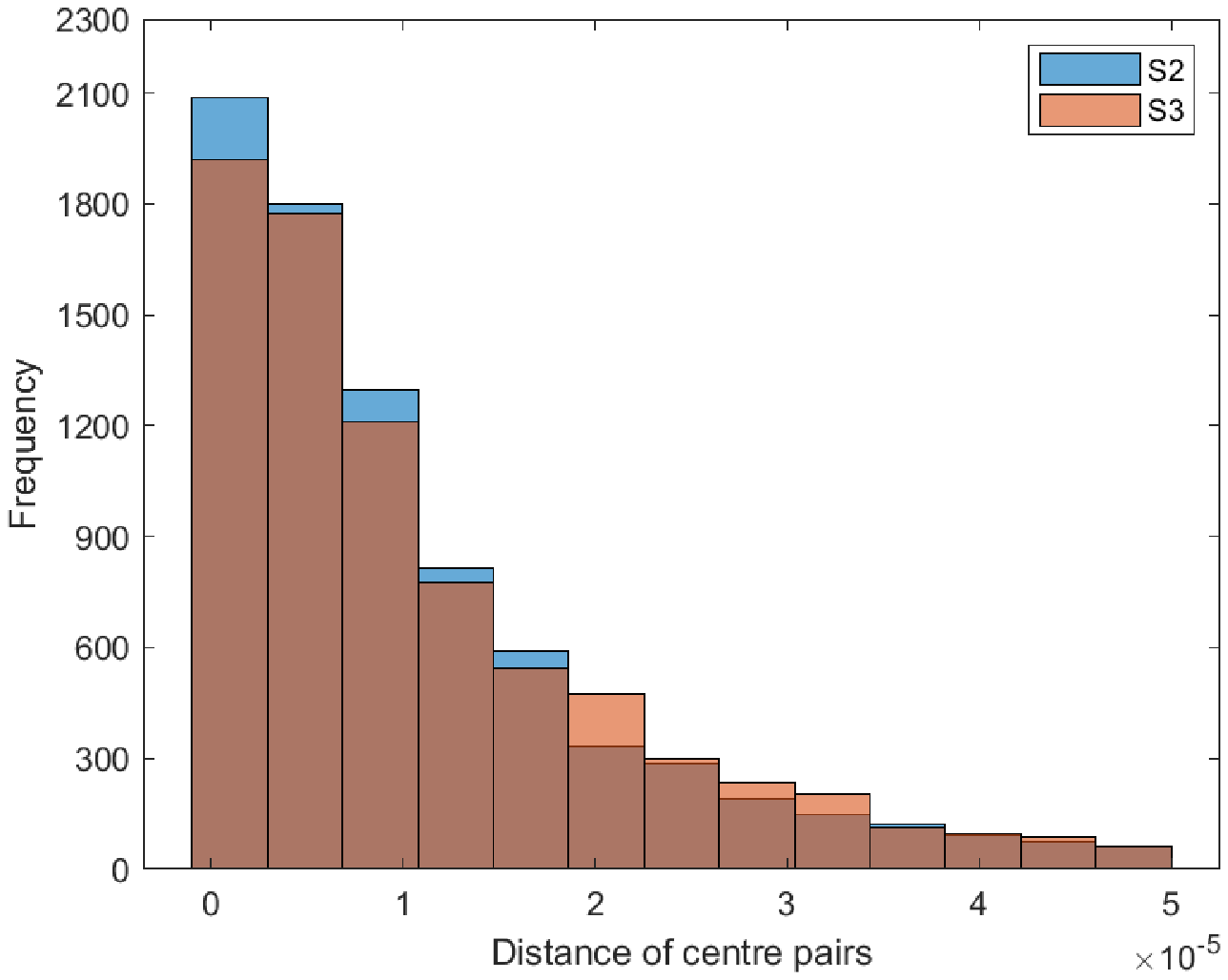}}
\end{minipage}
\caption{
For each class in VGGFace2, its corresponding nearest neighbour class can be found by comparing the positions of different class centres. (a), (b) and (c) show the the distributions of the distances between every class centre and its corresponding nearest class centre. Specifically, (a) shows the the distribution in the case of using the features generated by Scheme \uppercase\expandafter{\romannumeral2} (without using MML).  (b) shows the the distribution in the case of using the features generated by Scheme \uppercase\expandafter{\romannumeral3} (using MML). (c) shows the comparison results of (a) and (b), where S2 and S3 represent Scheme \uppercase\expandafter{\romannumeral2} and Scheme \uppercase\expandafter{\romannumeral3}, respectively.}
\label{fig:distribution_class_centres} %% label for entire figure
\end{figure*}

\subsection{Discussion}
\subsubsection{Whether MML can truly enlarge distances of the closest class centre pairs that are smaller than the specified minimum margin} To verify this point, we use the deep models trained by Scheme \uppercase\expandafter{\romannumeral2} (Softmax Loss + Centre Loss) and Scheme \uppercase\expandafter{\romannumeral3} (Softmax Loss + Centre Loss + MML) to extract the features of all the images from a cleaned version of VGGFace2 dataset \cite{cao2017vggface2}. The details of the cleaned dataset and the training process of these two models can be found in \ref{section:settings}. The difference between Scheme \uppercase\expandafter{\romannumeral2} and Scheme \uppercase\expandafter{\romannumeral3} is that Scheme \uppercase\expandafter{\romannumeral3} employs MML as a part of the supervision signal but Scheme \uppercase\expandafter{\romannumeral2} does not. With the extracted features, we calculate the centre position for each class and then calculate the distance between each class centre and its corresponding closest neighbour class centre. The distributions of the distances of these class centres are shown in Figure \ref{fig:distribution_class_centres}. Figure \ref{fig:h1} and Figure \ref{fig:h2} show the distance distributions of Scheme \uppercase\expandafter{\romannumeral2} and Scheme \uppercase\expandafter{\romannumeral3}, respectively. Figure \ref{fig:h3} makes a comparison between Scheme \uppercase\expandafter{\romannumeral2} and Scheme \uppercase\expandafter{\romannumeral3}, from which we can see that Scheme \uppercase\expandafter{\romannumeral3} has smaller values on the first five bins while owns larger values on the rest of the bins. This indicates that MML enlarges the distance of some neighbour centre pairs, therefore increases the quantity of the centre pairs having large margin.

\subsubsection{Whether MML can truly improve the performance of the model on face recognition} 
To answer this question, we conduct extensive experiments on different benchmark datasets as illustrated in Section \ref{section:Experiments}. The experimental types include face verification, face identification, image-based recognition and video-based recognition. Results show that the proposed method can beat the baseline methods as well as some state-of-the-art methods.

\section{Experiments}\label{section:Experiments}
In this section, we describe the implementation details of the experiments, investigate the influence of the parameters $\beta$ and $\mathcal{M}$, and evaluate the performance of the proposed method. The evaluations are conducted on MegaFace \cite{kemelmacher2016megaface}, FaceScrub \cite{ng2014data}, LFW \cite{LFWTech} and YTF \cite{wolf2011face} datasets with face identification and face verification tasks. Face identification and face verification are two main tasks of face recognition. Face verification aims at verifying whether two faces are from the person, answering `Yes' or `No', which is a binary classification problem. Face identification is to identifying the ID of a face, answering the exact ID, which is a multi-classification problem.

\subsection{Experiment Details} \label{section:settings}
\textbf{Training data.} In all experiments, we use VGGFace2 \cite{cao2017vggface2} as our training data. To ensure the reliability and the accuracy of the experimental results, we removed all the face images that might be overlapped with the benchmark datasets. As the label noise in the VGGFace2 is very low, no data cleaning has been applied. The final training dataset contains 3.05M face images from
8K identities.

\textbf{Data preprocessing.}
MTCNN \cite{zhang2016joint} is applied to all the face images for landmark location, face alignment and face detection. If face detection fails on a training image, we simply discard it; if it fails on a testing image, the provided landmarks are used instead. All the training and testing images are cropped to 160*160 RGB images. To augment the training data, we also perform random horizontal flipping on the training images. To improve the recognition accuracy, we concatenate the features of the original testing image and its horizontally flipped counterpart.

\textbf{Network settings.}
Based on Inception-ResNet-v1 \cite{DBLP:journals/corr/SzegedyIV16}, we implemented and trained three models by Tensorflow \cite{abadi2016tensorflow} according to three supervision schemes: Softmax Loss (Scheme \uppercase\expandafter{\romannumeral1}), Softmax Loss + Centre Loss (Scheme \uppercase\expandafter{\romannumeral2}), and Softmax Loss + Centre Loss + MML (Scheme \uppercase\expandafter{\romannumeral3}). We train these three models on one GPU (GTX 1080 Ti), and we set 90 as the batch size, 512 as the embedding size, 5e-4 as the weight decay and 0.4 as the keep probability of the fully connected layer. The total number of iterations is 275K, costing about 30 hours. The learning rate is initiated as 0.05 and is divided by 10 every 100K iterations. All three schemes use the same parameter settings except that Scheme \uppercase\expandafter{\romannumeral3} loads the trained model of Scheme \uppercase\expandafter{\romannumeral2} as the pre-trained model before training starts, as this way makes Scheme \uppercase\expandafter{\romannumeral3} achieves better recognition performance.

\textbf{Test settings.}
During the testing, we try our best to find the parameter settings that lead to highest performance. The $\alpha$ and $\beta$ in Eq.(\ref{eq:MML}) are set to be 5e-5 and 5e-8, respectively. The minimum margin of MML is set to be 280. The deep feature of each image is obtained from the output of the fully connected layer, and we concatenate the features of the original testing image and its horizontally flipped counterpart, therefore the resulting feature size of each image is $2*512$ dimensions. The final verification results are achieved by comparing the threshold with the Euclidean distance of two features

\subsection{Influence Analysis on Parameters $\beta$ and $\mathcal{M}$}
$\beta$ is the hyper-parameter for adjusting the impact of MML in the combination. $\mathcal{M}$ is the designated minimum margin. These two parameters influence the performance of the proposed method. Therefore, how to set these two parameters is a question worthy of study.

\begin{figure*}[!htbp]
\begin{minipage}{0.5\textwidth}
\subfigure[]{
\label{fig:subfig_Para_M} %% label for second subfigure
\includegraphics[width=3.4in,height=2.4in]{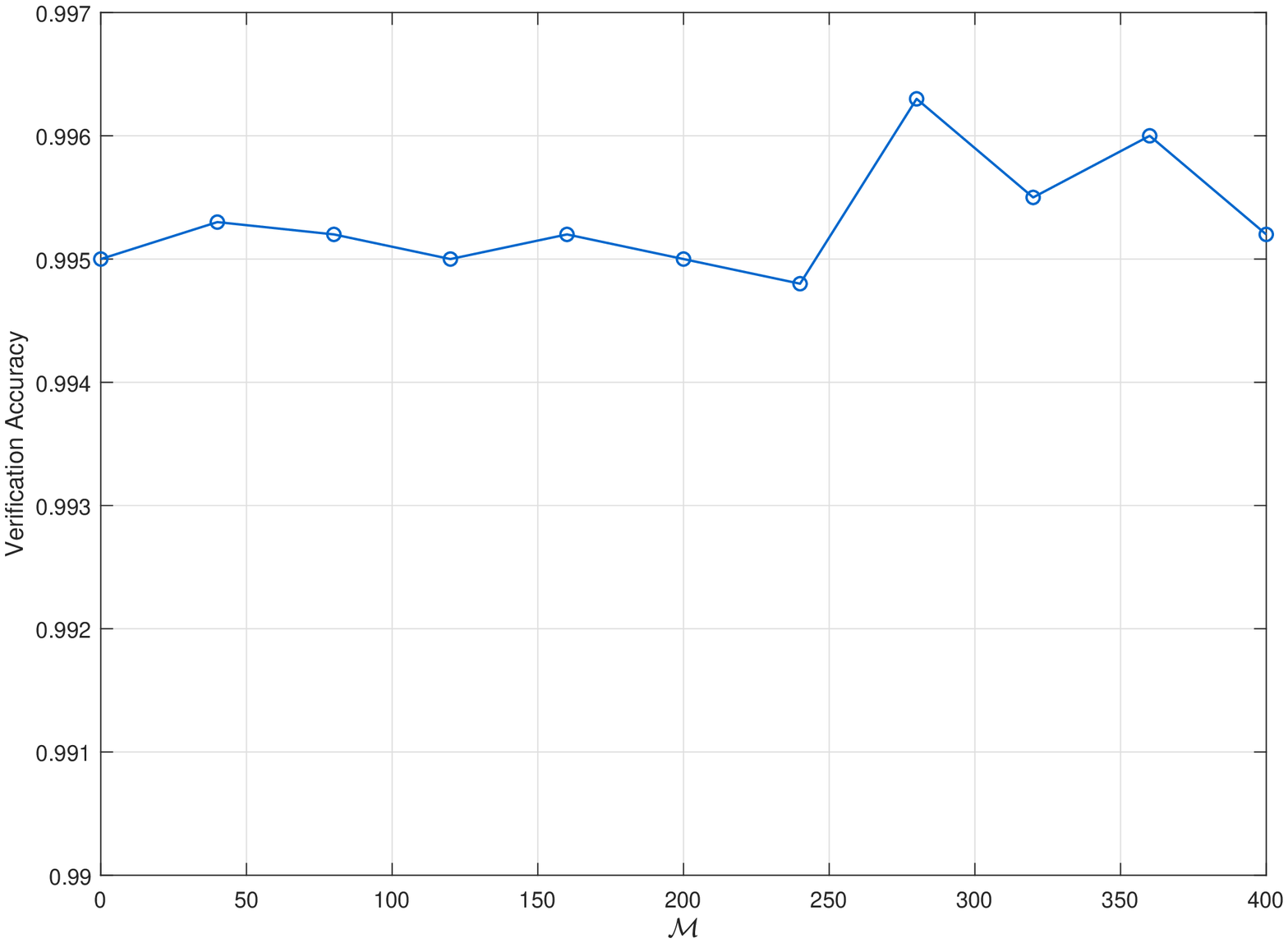}}
\end{minipage}
\begin{minipage}{0.5\textwidth} 
\hfill
\subfigure[]{
\label{fig:subfig_Para_beta} %% label for first subfigure
\includegraphics[width=3.4in,height=2.4in]{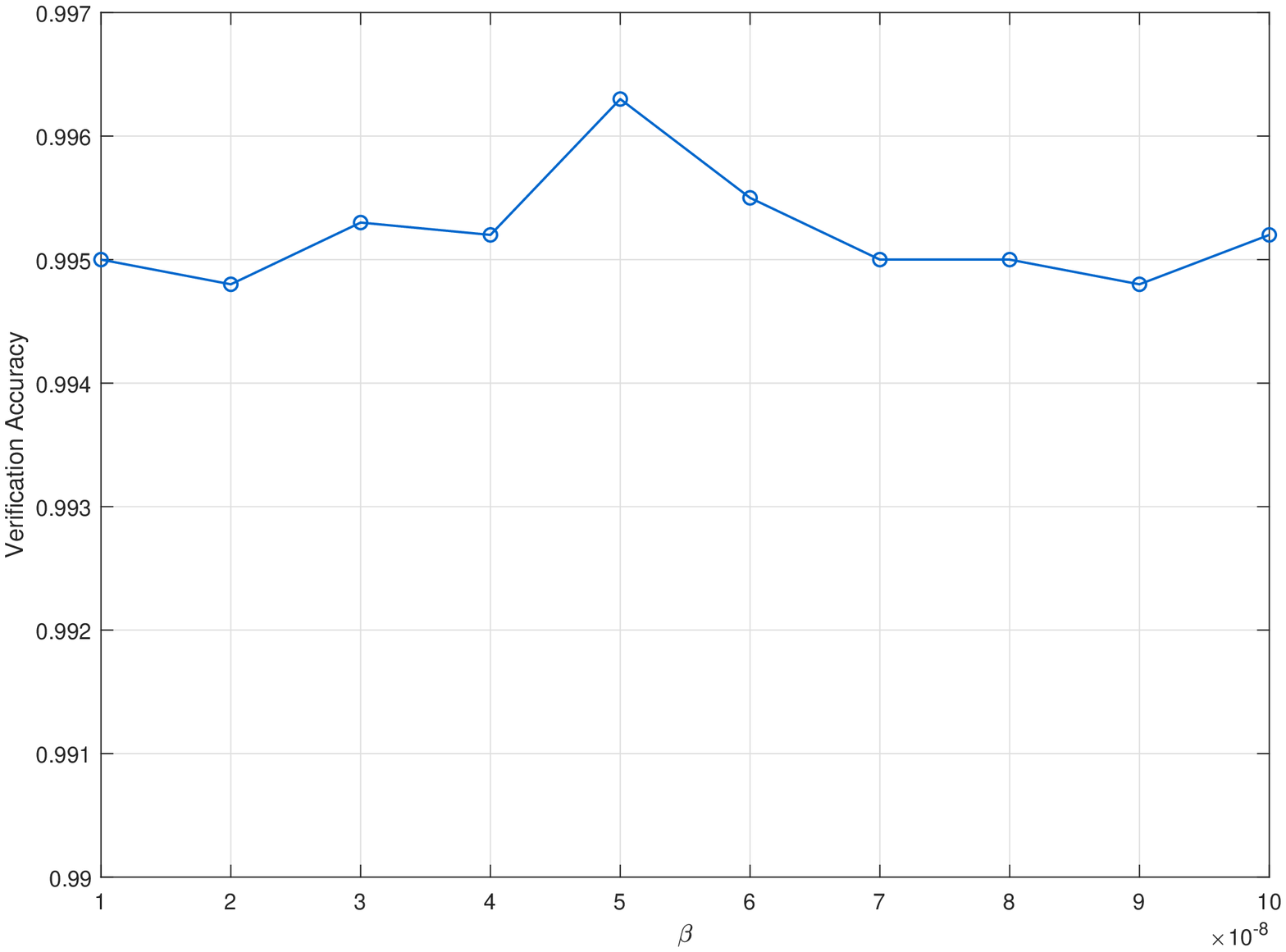}}
\end{minipage}
\caption{Face verification accuracies on LFW dataset with two groups of models: (a) fixed $\beta$ = 5e-8, and different $\mathcal{M}$, (b) fixed $\mathcal{M}$ = 280, and different $\beta$}.
\label{fig:Para_analysis} %% label for entire figure
\end{figure*}

We conduct two experiments on LFW dataset. In the first experiment, we fixed $\beta$ to 5e-8, and observe the influence of $\mathcal{M}$ on the verification performance as shown in Figure \ref{fig:subfig_Para_M}. In the second experiment, we fixed $\mathcal{M}$ to 280, and evaluate the relationship between $\beta$ and the verification accuracy as shown in Figure \ref{fig:subfig_Para_beta}. From Figure \ref{fig:subfig_Para_M}, we can see that setting $\mathcal{M}$ to 0, namely without using MML, is not proper, as it leads to low accuracy. The highest accuracy appears when $\mathcal{M}$ is 280. From Figure \ref{fig:subfig_Para_beta}, we can observe that the verification performance remains stable with a wide range of $\beta$, but reaches its peak value when $\beta$ is 5e-8. The above experimental results show that choosing the proper values for $\mathcal{M}$ and $\beta$ can improve the verification performance of the learned model. Therefore, in the subsequent experiments, we fixed $\mathcal{M}$ and $\beta$ to 280 and 5e-8, respectively.

\begin{figure*}[!htbp]
\begin{minipage}{0.5\textwidth} 
\subfigure[]{
\label{fig:subfig_CMC} %% label for second subfigure
\includegraphics[width=3.4in,height=2.4in]{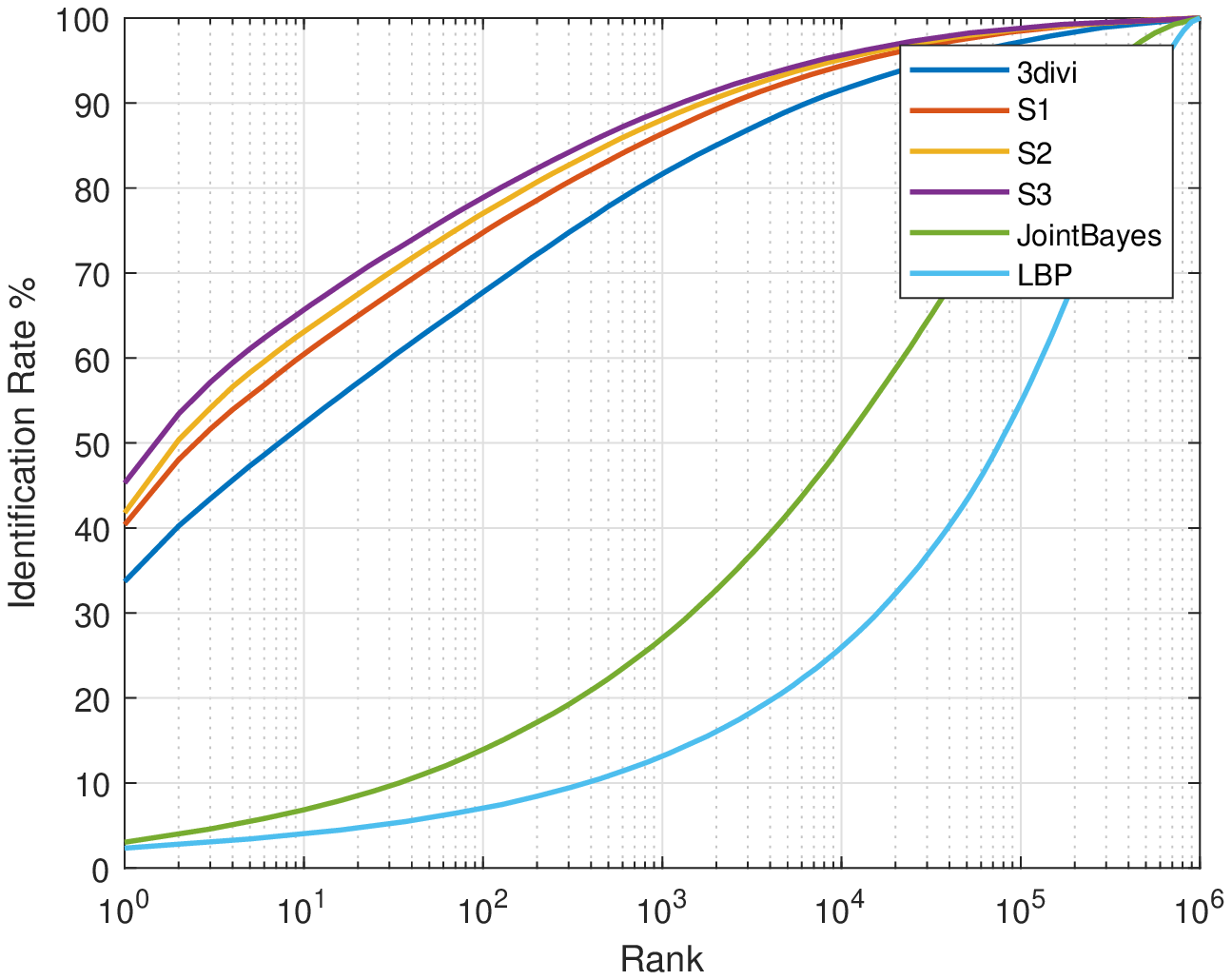}}
\end{minipage}
\begin{minipage}{0.5\textwidth} 
\hfill
\subfigure[]{
\label{fig:subfig_ROC} %% label for first subfigure
\includegraphics[width=3.4in,height=2.4in]{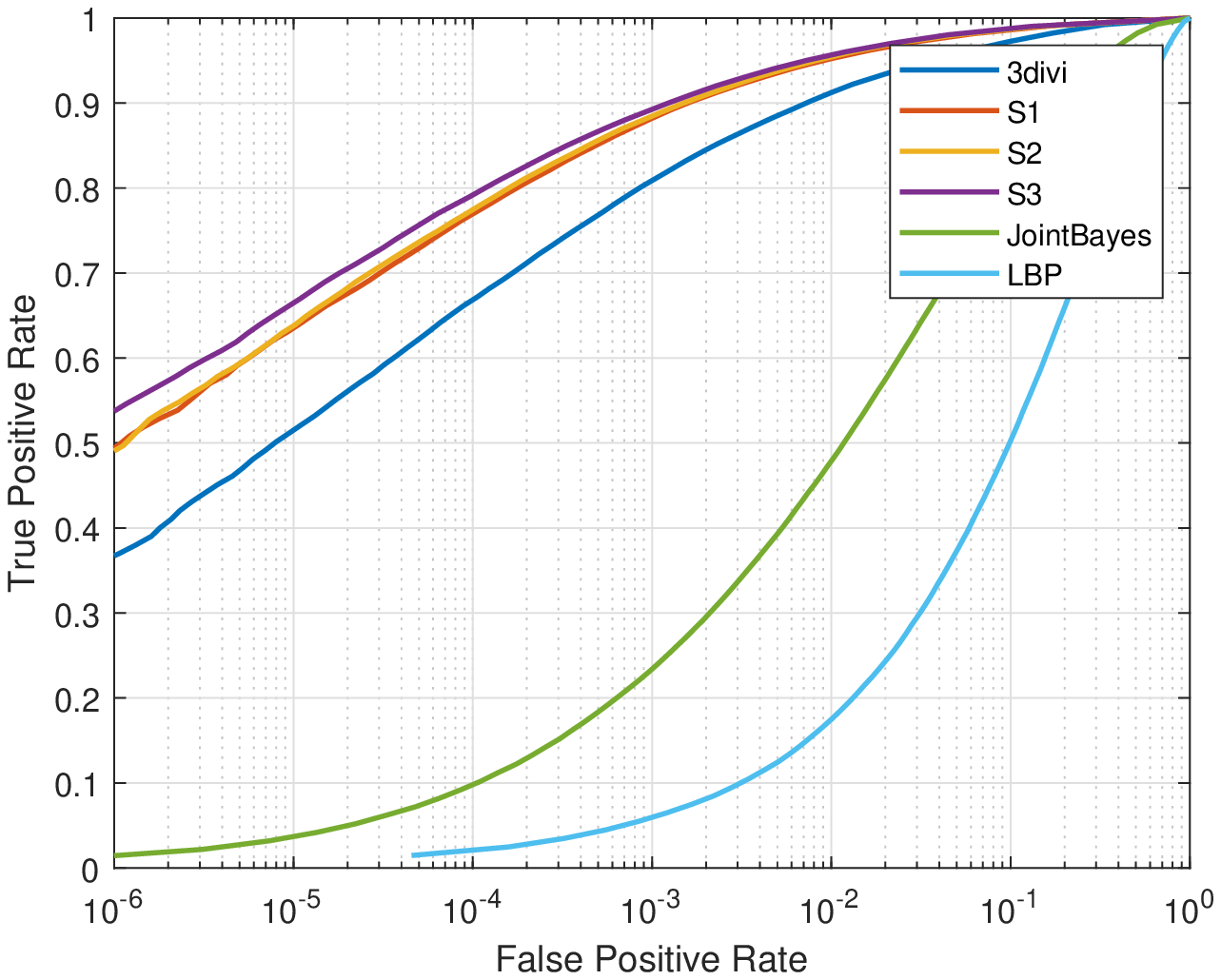}}
\end{minipage}
\caption{(a) reports the CMC curves of different methods with 1M distractors on MegaFace Set 1. (b) reports the ROC curves of different methods with 1M distractors on MegaFace Set 1. S1, S2 and S3 represent Scheme \uppercase\expandafter{\romannumeral1}, Scheme \uppercase\expandafter{\romannumeral2} and Scheme \uppercase\expandafter{\romannumeral3}, respectively. Apart from the S1 to S3, the results of other methods are provided by MegaFace team.}
\label{fig:CMC_ROC} %% label for entire figure
\end{figure*}

\begin{table*}[]
\centering
\caption{The identification rates and the verification rates of different methods on Megaface and FaceScrub datasets with 1M distractors.}
\label{Tabel:Megaface}
\begin{tabular}{l|c|c|c|c|c}
\hline
\multicolumn{1}{c|}{Methods} & Rank1@$10^6$ & Rank100@$10^6$ & VR@FAR$10^{-6}$ & VR@FAR$10^{-5}$ &VR@FAR $10^{-4}$ \\ \hline
3divi & 33.70\% & 67.74\% & 36.76\% & 51.56\% & 66.88\% \\
zJointBayes & 3.02\% & 13.94\% & 1.44\% & 3.70\% & 9.83\% \\
zLBP & 2.33\% & 7.04\% & - & - & 2.10\% \\ \hline
Scheme \uppercase\expandafter{\romannumeral1} & 41.39\% & 74.77\% & 49.46\% & 63.53\% & 76.97\% \\
Scheme \uppercase\expandafter{\romannumeral2} & 41.79\% & 77.03\% & 49.12\% & 63.79\% & 77.50\% \\
Scheme \uppercase\expandafter{\romannumeral3} (Proposed) & \textbf{45.30\%} & \textbf{78.89\%} & \textbf{53.80\%} & \textbf{66.54\%} & \textbf{79.23\%} \\ \hline
\end{tabular}
\end{table*}

\subsection{MegaFace Challenge 1 on FaceScrub}
In this section, we conduct experiment with the MegaFace dataset \cite{kemelmacher2016megaface} and the FaceScrub dataset \cite{ng2014data}. MegaFace dataset consists of a million faces and their respective bounding boxes obtained from Flickr (Yahoo's dataset). FaceScrub dataset is a publicly available dataset containing 0.1M images from 530 identities. According to the experimental protocol of MegaFace Challenge 1, the MegaFace dataset is used as the distractor set, while the FaceScrub dataset is used as the test set. The experiments are conducted with the provided code \cite{kemelmacher2016megaface}, which only evaluates our methods on one of the three sets of MegaFace (Set 1). More details about the experimental protocol can be found in \cite{kemelmacher2016megaface}.

We compare the proposed method (Scheme \uppercase\expandafter{\romannumeral3}) with some existing ones, including (a) 3divi (the deep model from 3DiVi Company), JointBayes \cite{chen2012bayesian} and LBP \cite{ahonen_face_2006}, and (b) our baseline methods (Scheme \uppercase\expandafter{\romannumeral1} and Scheme \uppercase\expandafter{\romannumeral2}). In the face identification experiments, the Cumulative Match Characteristics (CMC) curves \cite{phillips2003evaluation} are calculated to measure the ranking capabilities of different methods, as illustrated by Figure \ref{fig:subfig_CMC}). In the face verification experiments, we use the Receiver Operating Characteristic (ROC) curves to evaluate the different methods. The ROC curves plot the False Accept Rate (FAR) of a 1:1 matcher versus the False Reject Rate (FRR) of the matcher which are shown in Figure \ref{fig:subfig_ROC}. Table \ref{Tabel:Megaface} lists the numeric results of different methods on identification rates and the verification rates with 1M distractors.

From Figure \ref{fig:subfig_CMC}, Figure \ref{fig:subfig_ROC} and Table \ref{Tabel:Megaface}, we can observe that JointBayes and LBP perform poorly compared with other deep learning-based methods, and almost fail in excluding the distractors. The proposed Scheme \uppercase\expandafter{\romannumeral3} consistently outperforms Scheme \uppercase\expandafter{\romannumeral1}, Scheme \uppercase\expandafter{\romannumeral2} and 3divi, which confirms the effectiveness of the proposed method.

\begin{figure*}
\begin{center}
\includegraphics[width=6.3in,height=1.9in]{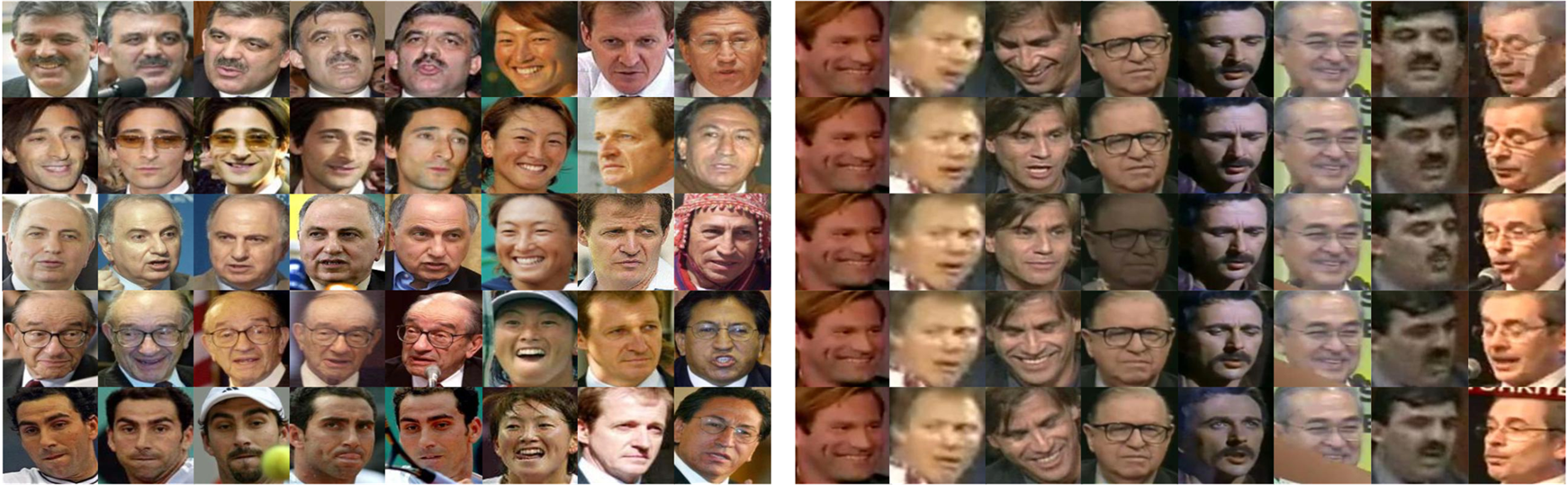}
\end{center}
\caption{Some examples from the LFW dataset (left) and the YTF dataset (right).}
\label{fig:LFW_YTF_samples}
\end{figure*}

\begin{table}[]
\centering
\caption{Verification performance of state-of-the-art methods on LFW and YTF datasets.}
\label{LFW-YTF-results}
\begin{tabular}{l|c|c|c}
\hline
\multicolumn{1}{c|}{Methods} & Images & LFW(\%) & YTF(\%) \\ \hline 
ICCV17' Range Loss \cite{zhang_range_2017} & 1.5M & 99.52 & 93.7 \\
CVPR17' Marginal Loss \cite{deng_marginal_2017} & 4M & 99.48 & 96.0 \\
CVPR15' DeepID2+ \cite{sun_deeply_2015} & & 99.47 & 93.2 \\
BMVC15' VGG Face \cite{parkhi_deep_2015} & 2.6M & 98.95 & 97.3 \\
CVPR14' Deep Face \cite{taigman_deepface:_2014} & 4M & 97.35 & 91.4 \\
CVPR15' Fusion \cite{taigman_web-scale_2015} & 500M & 98.37 & \\
ICCV15' FaceNet \cite{schroff2015facenet} & 200M & 99.63 & 95.1 \\
arXiv15' Baidu \cite{liu_targeting_2015} & 1.3M & 99.13 & \\
ECCV16' Centre Loss \cite{wen_discriminative_2016} & 0.7M & 99.28 & 94.9 \\
NIPS16' Multibatch \cite{tadmor_learning_2016} & 2.6M & 98.20 & \\
ECCV16' Aug \cite{masi_we_2016} & 0.5M & 98.06 & \\
arXiv18' ArcFace \cite{deng_arcface:_2018} & 7.1M & 99.83 & \\
CVPR17' SphereFace \cite{liu_sphereface:_2017} & 0.5M & 99.42 & 95.0 \\
arXiv18' CosFace \cite{wang_cosface:_2018} & 5M & 99.73 & 97.6 \\ \hline
Scheme \uppercase\expandafter{\romannumeral1} & 3.05M & 99.43 & 94.9 \\
Scheme \uppercase\expandafter{\romannumeral2} & 3.05M & 99.50 & 95.1 \\
Scheme \uppercase\expandafter{\romannumeral3} (Proposed) & 3.05M & 99.63 & 95.5 \\ \hline
\end{tabular}
\end{table}

\subsection{Comparison with the State-of-the-art Methods on LFW and YTF Datasets}

In this section, we evaluate the proposed method on two public benchmark datasets -- LFW \cite{LFWTech} and YTF \cite{wolf2011face} datasets according to the settings in Section \ref{section:settings}. Some preprocessed examples from these two datasets are shown in Figure \ref{fig:LFW_YTF_samples}.

LFW dataset is collected from the web, which contains 13,233 face images with large variations in facial paraphernalia, pose and expression. These face images come from 5749 different identities where 4069 of them have one image and the remaining 1680 identities have at least two images. LFW utilises the Viola-Jones face detector, which is the only constraint on the faces collected. We follow the standard experimental protocol of unrestricted with labelled outside data \cite{huang2014labeled} and test 6,000 face pairs according to the given pair list.

YTF dataset consists of 3,425 videos obtained from YouTube. These videos come from 1,595 identities with an average of 2.15 videos for each person. The frame number of the video clips ranges from 48 to 6,070, and the average is 181.3 frames. Also, we follow the standard experimental protocol of unrestricted with labelled outside data to evaluate the performance of the relevant methods on the given 5,000 video pairs.

Table \ref{LFW-YTF-results} shows the results of the proposed method and the state-of-the-art methods on LFW and YTF datasets, from which we can observe the followings. 
\begin{itemize}
\item The proposed Scheme III outperforms Scheme I (Softmax Loss only) and Scheme II (Softmax and Centre Loss), increasing the verification performance both on LFW and YTF datasets. On LFW, the accuracy improves from 99.43\% and 99.50\% to 99.63\%, while on YTF, the accuracy increases from 94.9\% and 95.1\% to 95.5\%. This demonstrates the effectiveness of the MML, also demonstrates the effectiveness of the combination of Softmax Loss + Centre Loss + MML.
\item Compared with the state-of-the-art methods, the proposed method has an accuracy higher than most of the methods on LFW and YTF datasets. Only ArcFace and CosFace slightly outperform the proposed method by 0.2\% and 0.1\% on LFW, however both of them utilise nearly double-sized training data and require much longer time and higher GPU memory on training. Using one GPU (GTX 1080 Ti, Memory size: 11GB), ArcFace needs at least 200 hours on training (the batch size is set to be 64 in order to fit the memory size of the GPU), while the proposed method only needs 30 hours for training. This shows the advantage of the whole framework including data preprocessing, network settings and MML.
\end{itemize}

\section{Conclusion}\label{Conclusion}
In this paper, a new loss function -- Minimum Margin Loss (MML) is presented to guide deep neural networks to learn highly discriminative face features. MML is aimed at enlarging the margin of those `unqualified' class centre pairs. To verify the effectiveness of the proposed method, extensive experiments are conducted. The results on the popular MegaFace, LFW and YTF datasets show that the proposed method has achieved the state-of-the-art performance.

\bibliographystyle{IEEEtran}

\end{document}